\documentclass[10pt,twocolumn,letterpaper]{article}

\usepackage{wacv}

\usepackage{algorithm}
\usepackage{algpseudocode}

\usepackage{placeins}

\definecolor{wacvblue}{rgb}{0.21,0.49,0.74}
\usepackage[pagebackref,breaklinks,colorlinks,allcolors=wacvblue]{hyperref}

\def\wacvPaperID{291}
\def\confName{WACV}
\def\confYear{2027}

\title{Joint Rigid Alignment and Morphable Reconstruction from Partial 3D Face Scans}

\author{
Eli A. Underwood\textsuperscript{1}
\quad Nicolas Rondan\textsuperscript{2}
\quad Angel V. Peterchev\textsuperscript{1}
\quad J. Matias Di Martino\textsuperscript{1,2}
\\
\textsuperscript{1}Duke University
\quad
\textsuperscript{2}Universidad Católica del Uruguay
}

\begin{document}
\maketitle

\begingroup
\renewcommand{\thefootnote}{}
\footnotetext{Research reported in this publication was supported by the National Institute of Mental Health of the National Institutes of Health under Award Number R01MH129733. The opinions expressed in this publication are those of the authors and do not necessarily reflect the views of the National Institutes of Health.}
\endgroup

\begin{abstract}
Aligning a partial, real-world 3{D} face scan to a canonical representation requires solving two coupled problems at once: estimating the rigid pose that brings the scan into a common frame, and completing the facial shape in regions the scan never observed. Existing pipelines typically decouple these steps, running a generic point-cloud registration front end followed by a separate non-rigid fitting or deformation stage. Building on the GeoTransformer superpoint-matching architecture, we propose a single network that predicts both a rigid pose and a shape-complete point cloud in one forward pass: a cross-attention regressor, conditioned on transformer point features and predicted coarse correspondences, maps a partial scan to per-patch coefficients of a {3D} morphable face model learned offline from a corpus of 8{,}000 identities. The resulting morphed template is then used, in the same forward pass, as the target for fine-grained Sinkhorn matching and rigid pose estimation, supervised purely by geometric alignment to ground-truth morphed geometry, so no hand-labeled coefficients are required. Against six classical and learned non-rigid registration baselines on held-out identities, our method leads the strongest ones by 3--4$\times$ on per-vertex error and baselines that collapse the head onto the visible region by more than an order of magnitude, and is simultaneously the most accurate and fastest method on ground-truth-free scans from two independent real-scan datasets, NoW and NPHM. Code will be made publicly available.
\end{abstract}

\section{Introduction}
\label{sec:intro}

{3D} scanners produce face geometry cheaply and at scale, but the resulting scans are rarely complete. A handheld or single-shot sensor typically observes one side of the face, misses the ears and much of the cranium, and returns a raw point cloud in an arbitrary, unknown pose. Turning such a scan into something usable for identity comparison, anthropometric measurement, prosthetic design, or animation retargeting requires solving two problems simultaneously. First, a \emph{rigid} problem: recovering the rotation and translation that bring the scan into a canonical reference frame. Second, a \emph{non-rigid} problem: completing the facial shape in the parts of the face the scan never saw, using prior knowledge of what faces look like, whether through a generic non-rigid solver or an iterative fit of a {3D} morphable face model against dense correspondences.

\begin{figure}[t]
  \centering
  \includegraphics[width=0.8\linewidth]{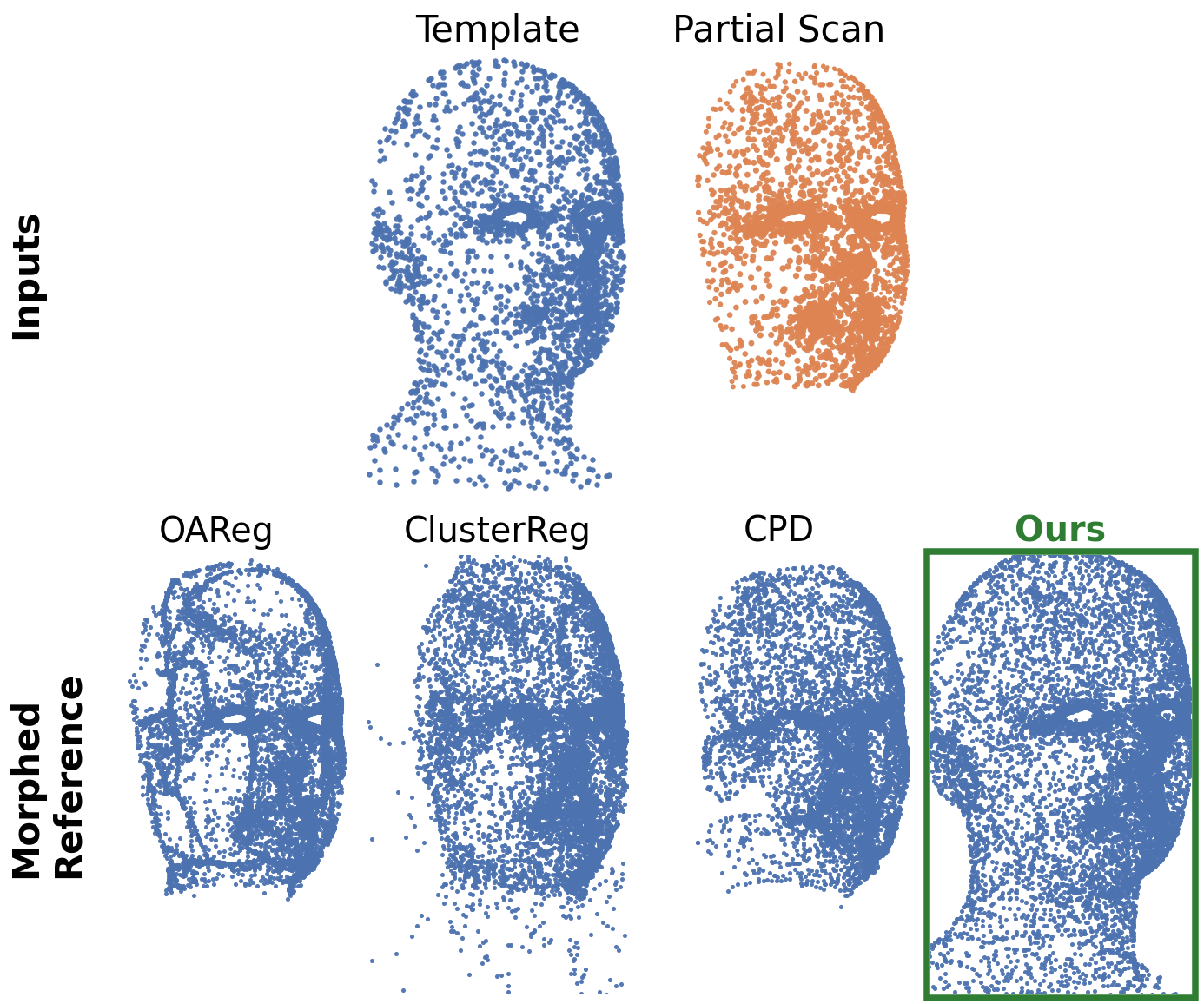}
  \caption{Mean-face template and partial validation scan (top), with the morphed references produced by three representative non-rigid registration baselines—OAReg~\cite{zhao2025oareg}, ClusterReg~\cite{zhao2024clusterreg}, and CPD~\cite{myronenko2010cpd}—and our method (bottom). The baselines collapse the head onto the visible region, whereas our method recovers a plausible full head from the same partial input.}
  \label{fig:teaser}
\end{figure}

We find this generic non-rigid prior is not just suboptimal but actively harmful on partial face scans. Benchmarking a representative set of non-rigid registration methods on partial views of a face database (\cref{sec:experiments}), the majority satisfy their smoothness regularizer by collapsing the head onto the visible region (\cref{fig:teaser}) rather than completing the occluded geometry, reaching per-vertex ground-truth error several times worse than methods that actually complete the shape.

\begin{figure*}[t]
  \centering
  \includegraphics[width=1.0\linewidth]{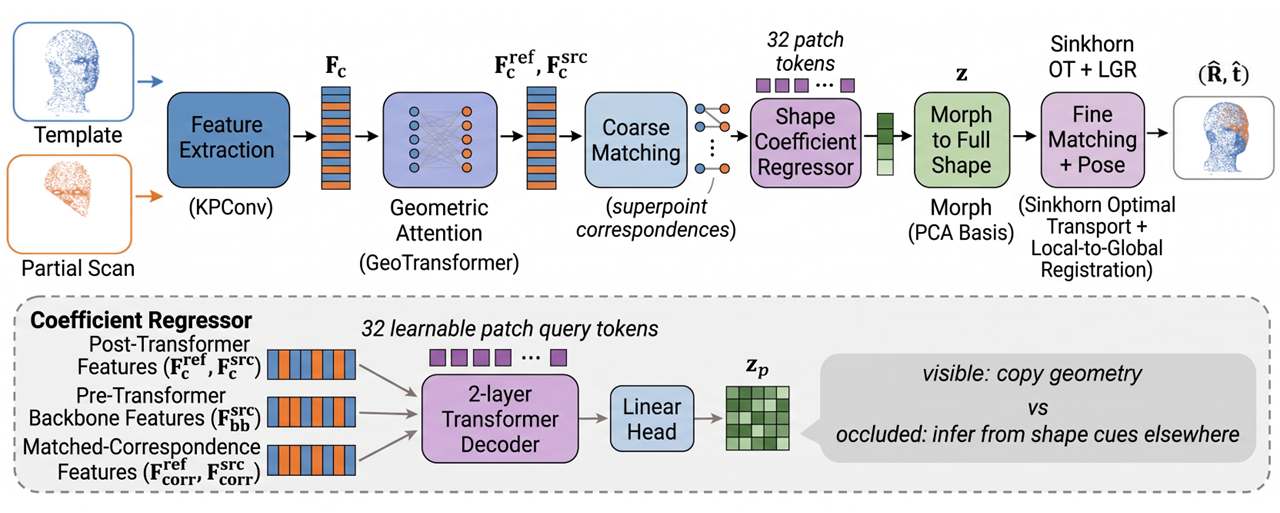}
  \caption{Pipeline overview. A shared KPConv (Kernel Point Convolution) backbone and GeoTransformer produce coarse superpoint correspondences between the mean-face template and a partial scan; a cross-attention regressor predicts per-patch morphable-basis coefficients $z$ (conditioned on post-transformer features, pre-transformer backbone features, and the coarse correspondences), which are decoded into a morphed template that updates the reference positions used by Sinkhorn optimal-transport (OT) fine matching and local-to-global registration (LGR) to produce the estimated pose $(\hat{R}, \hat{t})$. Bottom: detail of the coefficient regressor, showing its various input feature sources and the per-patch decoding into $z$.}
  \label{fig:pipeline}
\end{figure*}

Jointly predicting a rigid pose and a complete shape is not new, and neither is driving that completion with a {3D} morphable face model. What is new here is coupling the two through superpoint-level, nonlocal geometric matching, rather than treating rigid alignment and morphable-model fitting as two stages where one waits on the other. We condition the morphable-coefficient regressor directly on GeoTransformer's coarse superpoint features and predicted correspondences, so the same self- and cross-attention representation that establishes \emph{where} the scan overlaps the template also drives \emph{what} facial shape fills the rest of it, and gradients from the completion signal flow back into that matching stage itself. Specifically (\cref{fig:pipeline}), a KPConv~\cite{thomas2019kpconv} backbone and GeoTransformer~\cite{qin2022geotransformer} produce coarse superpoint correspondences between the scan and a fixed mean-face template. Then, a cross-attention regressor predicts per-patch coefficients of a {3D} morphable model~\cite{blanz1999morphable} from the pre- and post-transformer point features together with the predicted correspondences, deforming the template into a scan-specific, shape-complete morphed reference. This morphed reference is subsequently used, within the same forward pass, as the fine-matching target for Sinkhorn optimal transport~\cite{sinkhorn1967diagonal} and rigid pose estimation. Because the morphed template is a full, complete point cloud rather than a partial, noisy scan, morphing is supervised directly by dense geometric alignment to ground-truth morphed geometry, so no hand-labeled morphable-model coefficients are required. Gradients from this geometric signal flow back through the transformer and backbone, letting the coarse matching stage itself become shape-aware.

\paragraph{Contributions.}
\begin{enumerate}[noitemsep,topsep=0pt]
    \item An end-to-end architecture that jointly predicts a {3D} morphable face model's coefficients and rigid pose from a single partial scan in one forward pass, sharing geometric features between the coarse-matching, morphing, and fine-matching stages rather than treating them as independent modules. Because completion is conditioned on GeoTransformer's superpoint correspondences, the network localizes and adapts to whatever geometry is actually missing on a per-scan basis, while the same superpoint-level matching keeps global rigid alignment robust.
    \item A cross-attention morphable-coefficient regressor, conditioned on pre-transformer backbone features, post-transformer matched features, and the predicted coarse correspondences, trained with a dense geometric loss against ground-truth morphed geometry rather than hand-labeled coefficients.
    \item A short, scan-referenced flow-field refinement, warm-started from our end-to-end prediction, that closes the remaining surface-fit gap in the visible region while leaving occluded geometry governed by the learned prior.
    \item A rigorous evaluation protocol pairing per-vertex ground-truth error with Chamfer distance on a held-out validation set, exposing a critical collapse failure mode in standard non-rigid registration baselines that Chamfer distance alone conceals. We extend this evaluation to two externally sourced real-scan datasets, the Not-quite-in-the-Wild (NoW) benchmark~\cite{sanyal2019now} and a subset from the Neural Parametric Head Models (NPHM)~\cite{giebenhain2023nphm}. Across all settings, our method avoids the collapse failure and outperforms all baselines without external rigid alignment.

\end{enumerate}

\section{Related work}
\label{sec:related}

\paragraph{Rigid point cloud registration.}
Classical rigid registration alternates correspondence estimation and closed-form pose solving, from Iterative Closest Point (ICP)~\cite{besl1992icp} to feature-based Random Sample Consensus (RANSAC)~\cite{fischler1981ransac} pipelines built on descriptors such as Fast Point Feature Histograms (FPFH)~\cite{rusu2009fpfh}. Learned methods replace one or both stages with neural correspondence estimation: Predator~\cite{huang2021predator} learns overlap-aware keypoint detection for low-overlap scenes, while GeoTransformer~\cite{qin2022geotransformer} learns geometric self- and cross-attention over superpoints so accurate correspondences can be found without a RANSAC step. Dual Focus-Attention Transformer (DFAT)~\cite{fu2025dfat} focuses attention within local neighborhoods at both coarse and fine scales and Regor~\cite{zhao2025regor} regenerates rather than merely filters correspondences under high outlier ratios. 

\paragraph{Non-rigid registration and surface fitting.}
Non-rigid registration extends the rigid setting with a per-point or per-node deformation field, typically regularized for local smoothness or as-rigid-as-possible behavior. Optimal-step non-rigid ICP (NRICP)~\cite{amberg2007optimal} and Coherent Point Drift (CPD)~\cite{myronenko2010cpd} are foundational classical approaches. More recent work replaces the deformation model or correspondence step with learned components. GraphSCNet~\cite{qin2023graphscnet} prunes outlier correspondences using local spatial consistency over a deformation graph before solving with embedded deformation. Test-time-optimization methods forgo learned priors entirely: Neural Scene Flow Prior (NSFP)~\cite{li2021nsfp} fits an unregularized flow MLP against a Chamfer objective, OAReg~\cite{zhao2025oareg} adds an occlusion-aware correntropy term and a local linear-rigidity regularizer, and ClusterReg~\cite{zhao2024clusterreg} reformulates registration as unsupervised clustering with a closed-form displacement field. None encode a prior over what a face specifically looks like, as their regularizers constrain local rigidity or smoothness, not global plausibility; \cref{sec:experiments} shows this matters, since smoothness or rigidity regularization alone does not prevent global shape collapse on partial face scans. Our refinement (\cref{sec:method:refine}) reuses Neural Scene Flow Prior's~\cite{li2021nsfp} unregularized, per-scan flow-field optimizer directly, but warm-started from our learned prediction rather than a generic initialization, improving accuracy with far fewer steps to converge (\cref{sec:experiments:realscan}).
\paragraph{{3D} morphable face models.}
3D morphable models (3DMMs) represent face and head shape as a low-dimensional basis fit to a corpus of registered scans, from the original 3DMM~\cite{blanz1999morphable} and the Basel Face Model~\cite{paysan2009basel} to expressive models such as FLAME~\cite{li2017flame} and large-scale face-and-head models~\cite{ploumpis2019combining}. These are most commonly fit to an input via iterative optimization, or regressed from a single RGB image by a learned encoder~\cite{sanyal2019now}, typically assuming a roughly frontal view and a known pose. Our regressor instead operates on {3D} point features and jointly recovers the unknown $SE(3)$ pose alongside completion. We instead use a morphable model as the target representation for a registration network. Concretely, the decoder regresses local, patch-wise coefficients directly from geometric point features, and the morphed points are used, in the same forward pass, as the target for classical fine-grained point matching, coupling morphable-model fitting to registration rather than treating it as a downstream step. 
\paragraph{Learned shape completion priors.}
Generic point-cloud completion networks~\cite{yuan2018pcn,yu2021pointr} learn to hallucinate missing geometry from partial input, typically trained per-category on synthetic shape collections such as ShapeNet. Our regressor is a domain-specialized instance of this idea: rather than a generic latent shape code, the completion target is constrained to the linear span of a face-specific PCA basis, which regularizes the output to always be a plausible face and makes the completion interpretable as anatomically localized coefficients.
\Cref{sec:experiments} shows this domain-specific prior is precisely what separates methods that complete occluded facial geometry from methods that merely avoid destroying the visible region.

\section{Method}
\label{sec:method}

Given a partial, arbitrarily-posed {3D} point cloud $\mathbf{S} \in \mathbb{R}^{N_s \times 3}$ of a face ($N_s$ vertices), our goal is to jointly recover (i) the rigid transform $(R, t) \in SE(3)$ that aligns $\mathbf{S}$ to a canonical frame, and (ii) a shape-complete point cloud $\hat{\mathbf{M}}$, in the same fixed topology as a mean-face template $\bar{\mathbf{M}} \in \mathbb{R}^{N_v \times 3}$ ($N_v$ vertices), representing the scanned individual's full facial geometry. \Cref{fig:pipeline} gives an overview. Both the rigid pose $(R, t)$ and the completed point cloud $\hat{\mathbf{M}}$ come from a single network in one forward pass, conditioned on the same geometric point features, rather than independent rigid and non-rigid stages.

\subsection{Preliminaries: a collection of local morphable patch models}
\label{sec:method:prelim}
Our shape-completion signal comes from a linear morphable basis, fit offline to a corpus of $N{=}8{,}000$ synthetic head identities from the Universal Head Model (UHM) model~\cite{ploumpis2020uhm,foti2022swapdisentangled} sharing the mean template's fixed topology ($N_v{=}10{,}788$ vertices). Rather than fit one global basis over the whole head, which would let a coefficient's effect spread across the entire face and make partial, patch-local evidence hard to exploit, we build $P{=}32$ independent local models, one per overlapping patch of $\bar{\mathbf{M}}$. This decomposition is what makes the coarse superpoint correspondences from \cref{sec:method:backbone}'s GeoTransformer usable as completion evidence in the first place: because each patch has its own basis, \cref{sec:method:regressor}'s regressor can localize completion to the parts of the facial shape that a given patch's matched superpoints actually constrain, rather than entangling every prediction with the whole head at once. Patch $p$ is the set of $K_n{=}700$ nearest template vertices to a fixed anchor, so a vertex may belong to several patches. Let $\mathbf{x}_p^{(n)} \in \mathbb{R}^{3K_n}$ be the flattened positions of patch $p$'s vertices in corpus identity $n$. We compute the patch mean $\mu_p = \frac{1}{N}\sum_n \mathbf{x}_p^{(n)}$ and, by PCA over $\{\mathbf{x}_p^{(n)} - \mu_p\}_n$, an orthonormal basis $\mathbf{B}_p \in \mathbb{R}^{K \times 3K_n}$ of the leading $K$ directions of shape variation ($K{=}36$; \cref{tab:config}). A patch is reconstructed from coefficients $z_p \in \mathbb{R}^K$ by
\begin{equation}
  \hat{\mathbf{x}}_p = \mu_p + z_p^{\top} \mathbf{B}_p ,
  \label{eq:patch-recon}
\end{equation}
reshaped to $K_n{\times}3$ vertex positions. Assembling all $P$ patches into a full point cloud $g(z) \in \mathbb{R}^{N_v \times 3}$ is a fixed, deterministic gather with no learned parameters. Each vertex takes its position from a designated covering patch. $g$ is linear in $z$, so predicting the facial shape in occluded and visible regions alike reduces to predicting $z \in \mathbb{R}^{P \times K}$, and \cref{sec:method:regressor} describes the network that regresses $z$ from a partial scan. \Cref{alg:forward} summarizes how this basis, the regressor, and the morphing step of \cref{sec:method:morph} fit together in the full forward pass.

\begin{algorithm}[t]
\small
\caption{Forward pass: scan $\rightarrow$ $(\hat{\mathbf{M}}, \hat{R}, \hat{t})$}
\label{alg:forward}
\begin{algorithmic}[1]
\Require partial scan $\mathbf{S}$, template $\bar{\mathbf{M}}$
\State Extract backbone + GeoTransformer features for $\bar{\mathbf{M}}$ and $\mathbf{S}$ \Comment{\cref{sec:method:backbone}}
\State Coarsely match superpoints between $\bar{\mathbf{M}}$ and $\mathbf{S}$ \Comment{\cref{sec:method:backbone}}
\State Regress per-patch coefficients $z$ from matched features \Comment{\cref{sec:method:regressor}}
\State Reconstruct full point cloud $\hat{\mathbf{M}} = g(z)$ \Comment{\cref{eq:patch-recon}}
\State Update reference positions at all backbone levels to $\hat{\mathbf{M}}$ \Comment{\cref{sec:method:morph}}
\State Fine-match via Sinkhorn OT + LGR on $\hat{\mathbf{M}}$ to get $(\hat{R}, \hat{t})$ \Comment{\cref{sec:method:morph}}
\State \Return $\hat{\mathbf{M}}, \hat{R}, \hat{t}$
\end{algorithmic}
\end{algorithm}

\subsection{Backbone and coarse matching}
\label{sec:method:backbone}
We build a $k$-nearest-neighbor graph jointly over the raw mean-face template $\bar{\mathbf{M}}$ and the raw source scan $\mathbf{S}$ and encode both with a shared KPConv feature-pyramid backbone~\cite{thomas2019kpconv} at $L{=}4$ resolution levels, giving per-point features at a fine (Stage-1) and coarse (Stage-4) resolution for both. The coarsest superpoint features are refined by a GeoTransformer~\cite{qin2022geotransformer}, applying geometric self- and cross-attention (pairwise distances, embedded with Gaussian bandwidth $\sigma_d$, and triplet angles, both invariant to the unknown rigid pose) within and between the point sets, producing post-transformer features $\mathbf{F}^{\text{ref}}_c, \mathbf{F}^{\text{src}}_c$. A differentiable superpoint matching module then produces coarse correspondences from these features, supervised by a circle loss~\cite{sun2020circleloss} against ground-truth overlap (\cref{sec:method:losses}).

\subsection{Morphable-coefficient regression}
\label{sec:method:regressor}
Coarse correspondences alone identify \emph{where} the scan overlaps the template, but they say nothing about the occluded facial shape. We introduce a cross-attention regressor that predicts this separately, holding $P$ learnable patch query tokens that cross-attend, via a 2-layer Transformer decoder, to a memory built from three feature sources: the post-transformer template and scan superpoint features ($\mathbf{F}^{\text{ref}}_c, \mathbf{F}^{\text{src}}_c$), the pre-transformer scan backbone features $\mathbf{F}^{\text{src}}_{bb}$ (raw local geometry, before it is entangled with the template), and the features at the predicted coarse correspondences $\mathbf{F}^{\text{ref}}_{\text{corr}}, \mathbf{F}^{\text{src}}_{\text{corr}}$ (which parts of the scan are actually observed). A linear head maps each updated patch token to coefficients $z_p \in \mathbb{R}^{K}$ (\cref{sec:method:prelim}). Conditioning on both pre- and post-transformer features alongside the explicit correspondence set lets the regressor decouple direct geometric transfer in visible regions from learned completion in occluded areas.

\subsection{Morphing the reference and fine matching}
\label{sec:method:morph}
The predicted coefficients reconstruct a morphed template $\hat{\mathbf{M}} = g(z) \in \mathbb{R}^{N_v \times 3}$ (\cref{sec:method:prelim,sec:method:regressor}). Because the pooled backbone points at each stage are exact copies of vertices of $\bar{\mathbf{M}}$, but KPConv pooling does not retain which vertex index each pooled point came from, we recover this correspondence via an exact-match nearest-neighbor lookup against the ordered vertex set of $\bar{\mathbf{M}}$, yielding a fixed index for every coarse- and fine-level template point. Since $\hat{\mathbf{M}}$ preserves the same global vertex ordering as $\bar{\mathbf{M}}$, we use these indices to directly gather the morphed positions from $\hat{\mathbf{M}}$, so the morphed geometry, rather than the mean face, is what fine matching operates on. Ground-truth coarse correspondences for the circle loss are likewise computed \emph{after} morphing. Fine matching then follows GeoTransformer's local-to-global registration~\cite{qin2022geotransformer}: point-to-point scores within each coarse-matched patch pair are normalized with a differentiable Sinkhorn optimal-transport layer~\cite{sinkhorn1967diagonal}, dense correspondences are extracted, and a robust local-to-global procedure aggregates per-patch rigid hypotheses into $(\hat{R}, \hat{t})$. The rigid pose is thus estimated against a target that already reflects the network's best guess at the scanned individual's facial shape, not just the population mean.

\subsection{Losses}
\label{sec:method:losses}
The network is trained with a sum of three terms: a circle loss on coarse correspondences~\cite{sun2020circleloss}, a matching loss on the fine-level Sinkhorn scores (positive pairs pulled toward their ground-truth match, unmatched rows/columns pulled toward the dustbin), and a morph supervision term. Our generated training pairs (\cref{sec:method:data}) provide full ground-truth morphed geometry, not just a partial, noisy scan, so the morph term is a direct, dense per-vertex loss,
\begin{equation}
  \mathcal{L}_{\text{morph}} = \frac{1}{N_v}\sum_{i=1}^{N_v} \left\lVert \hat{\mathbf{M}}_i - \mathbf{M}^{\text{gt}}_i \right\rVert_2,
  \label{eq:vertex-loss}
\end{equation}
where $\mathbf{M}^{\text{gt}} = g(z^{\text{gt}})$ is the full ground-truth morphed template, weighted ten times relative to the coarse and fine losses, dominating the training signal for the decoder. This is a geometric loss rather than a coefficient-regression loss. Because $z$ is never supervised directly, gradients flow from a shape-level alignment signal back through \cref{eq:vertex-loss} into the regressor and, because the memory it attends to includes post-transformer features, into the transformer and backbone as well.

\subsection{Training data}
\label{sec:method:data}
Ground-truth coefficients and rigid transforms are not available for scans outside our own generated corpus, so we train entirely on pairs generated from the corpus of \cref{sec:method:prelim}. The reference is the fixed mean face and per-patch bases computed over the full 8{,}000-identity corpus. Of those 8{,}000, 7{,}500 generate training pairs (2 views each) and 250 are held out for evaluation (\cref{sec:experiments:data}); the remaining 250 contribute only to the shared mean and per-patch bases, not to any pair. The source is a partial view of one identity, drawn with equal probability from one of two crop modes: \textbf{Face-Cut}, an axis-aligned crop with randomized bounds, or \textbf{Self-Occlusion}, a hidden-point-removal crop of the full head from a random viewpoint. A random $SO(3)$ rotation and translation are applied to either crop mode, simulating an arbitrarily-posed handheld or single-shot scan. At train time only, we apply additive Gaussian noise ($\pm$0.005 per axis), a random subsample (keep ratio $\sim\mathcal{U}(0.5,1.0)$), and a voxel-density normalization pass to better match real scanner density irregularity. Ground-truth patch coefficients $z^{\text{gt}}$ are precomputed once per identity. The ground-truth transform is the actual $(R, t)$ needed to bring the source crop back into the reference frame. \Cref{tab:config} summarizes the resulting configuration.

\begin{table}[t]
  \centering
  \footnotesize
  \begin{tabular}{@{}ll@{}}
    \toprule
    Component & Value \\
    \midrule
    Backbone & voxel 0.025, 4 stages \\
    Transformer $\sigma_d$ (train) & 0.16 \\
    Coarse targets / correspondences & 128 / 256 \\
    Coarse / fine matching radius & 0.05 / 0.1 \\
    Sinkhorn iterations & 100 \\
    Morphable patches $\times$ components & $32 \times 36$ \\
    Loss weight (morph : coarse : fine) & $10 : 1 : 1$ \\
    Optimizer & Adam, lr $10^{-4}$, $\times0.95$/10 ep. \\
    Weight decay / grad.\ clip & $10^{-4}$ (excl.\ regressor) / 1.0 \\
    Epochs / batch size & 500 / 8$\times$4 GPUs (eff.\ 32) \\
    Noise / subsample aug.\ & $\pm$0.005 / keep $\sim\mathcal{U}(0.5,1)$ \\
    Voxel-density aug.\ & dedup $v \in \{0.03,\dots,0.06\}$ \\
    \bottomrule
  \end{tabular}
  \caption{Key configuration for the checkpoint used in all reported results. At real-scan (NoW and NPHM) inference only, $\sigma_d$ is scaled by a fixed $\times1.2$ correction factor.}
  \label{tab:config}
\end{table}
\begin{table*}[t]
  \centering
  \footnotesize
  \setlength{\tabcolsep}{11pt}
  \begin{tabular}{@{}lrrrrr@{}}
    \toprule
    & \multicolumn{2}{c}{GT err (mm) $\downarrow$}
    & \multicolumn{2}{c}{CD (mm) $\downarrow$}
    & \\
    Method & Self-Occlusion & Face-Cut & Self-Occlusion & Face-Cut & Time (s) $\downarrow$ \\
    \midrule
    NRICP~\cite{amberg2007optimal}
      & $31.86 \,\pm\, 16.56$
      & $29.62 \,\pm\, 10.36$
      & $3.87 \,\pm\, 9.27$
      & $1.71 \,\pm\, 6.63$
      & 64.45 \\
    CPD~\cite{myronenko2010cpd}
      & $46.65 \,\pm\, 16.51$
      & $44.97 \,\pm\, 10.16$
      & $1.55 \,\pm\, 0.19$
      & $1.47 \,\pm\, 0.15$
      & 39.95 \\
    NSFP~\cite{li2021nsfp}
      & $9.99 \,\pm\, 4.27$
      & $8.88 \,\pm\, 3.92$
      & $0.50 \,\pm\, 0.25$
      & $0.31 \,\pm\, 0.20$
      & 15.40 \\
    GraphSCNet~\cite{qin2023graphscnet}
      & $8.56 \,\pm\, 3.75$
      & $7.77 \,\pm\, 3.63$
      & $2.06 \,\pm\, 1.42$
      & $1.35 \,\pm\, 0.82$
      & 40.60 \\
    ClusterReg~\cite{zhao2024clusterreg}
      & $42.71 \,\pm\, 17.03$
      & $38.22 \,\pm\, 11.02$
      & $0.87 \,\pm\, 0.28$
      & $0.91 \,\pm\, 0.18$
      & 41.40 \\
    OAReg~\cite{zhao2025oareg}
      & $36.67 \,\pm\, 11.63$
      & $34.92 \,\pm\, 7.89$
      & $2.15 \,\pm\, 0.88$
      & $1.28 \,\pm\, 0.64$
      & 14.25 \\
    Ours
      & $\mathbf{2.42 \,\pm\, 1.05}$
      & $\mathbf{2.12 \,\pm\, 0.67}$
      & $\mathbf{0.08 \,\pm\, 0.04}$
      & $\mathbf{0.08 \,\pm\, 0.02}$
      & \textbf{2.53} \\
    \bottomrule
  \end{tabular}

  \caption{Val set (250 identities $\times$ 2 views = 500 pairs: 266 Self-Occlusion / 234 Face-Cut pairs), GT vertex error and 1-way CD (partial scan$\to$deformed-template, visible region only, same definition as \cref{tab:now}), median $\pm$ std over pairs, all methods timed on the same hardware. Ours leads every column. Four of six baselines (NRICP, CPD, ClusterReg, OAReg) have GT error an order of magnitude worse than Ours despite ClusterReg posting some of the best CD in the table alongside 38--47mm GT error: CD's blind spot to implausible completions, visible in single rows (\cref{sec:experiments:metrics}).}
  \label{tab:val}
\end{table*}

\subsection{Scan-referenced refinement}
\label{sec:method:refine}
The forward pass above (\cref{sec:method:backbone,sec:method:regressor,sec:method:morph}) never looks at the scan surface directly when producing $\hat{\mathbf{M}}$: $\hat{\mathbf{M}}$ is a function of learned coarse features, constrained to the morphable-model basis, and is never pulled toward the scan by any per-point data term. This is exactly what keeps it from the collapse failure of smoothness-regularized baselines (\cref{sec:experiments:results}), but it also means $\hat{\mathbf{M}}$ does not exploit fine, scan-specific surface detail that \emph{is} available in the visible region and that a direct scan-fitting objective would recover.

Our pipeline therefore closes with a short test-time refinement that recovers this detail without giving up the completion above. It takes the forward pass's rigid pose $(\hat R, \hat t)$ and morphed template $\hat{\mathbf{M}}$ as initialization for a short run of a NSFP-style~\cite{li2021nsfp} per-scan flow-field optimization against the raw scan, the same unregularized flow-MLP objective NSFP uses from scratch, but warm-started from our prediction instead of the population mean and an ICP-derived pose. Because the flow field is fit against the observed scan, it can only meaningfully move points the scan actually constrains. In the occluded region, where no scan data exists to pull it anywhere, the geometry stays close to $\hat{\mathbf{M}}$, so the refinement inherits our shape-completion behavior there rather than overriding it. Unlike running NSFP from scratch, this refinement never pays for an external rigid-ICP pre-alignment stage (\cref{sec:experiments:baselines}), so far fewer optimization steps are needed to converge. \cref{sec:experiments:ablations} isolates the forward pass alone.

\section{Experiments}
\label{sec:experiments}

\subsection{Datasets and protocol}
\label{sec:experiments:data}
We evaluate on three datasets: our own synthetic validation set, with ground truth, and two independent, external real-scan datasets, without ground truth. \textbf{Validation} (val) is drawn from a 250-identity pool of UHM-sampled heads~\cite{ploumpis2020uhm,foti2022swapdisentangled} held out from training, each identity rendered with two partial views (500 pairs total: 266 Self-Occlusion / 234 Face-Cut, grouped by the crop mode actually sampled for each view, \cref{sec:method:data}), each paired with the corresponding full ground-truth morphed mesh, enabling direct, alignment-independent per-vertex accuracy (\cref{sec:experiments:metrics}). We report Self-Occlusion and Face-Cut separately, since occlusion crops have systematically more missing geometry. \textbf{NoW} is the test split of the NoW benchmark~\cite{sanyal2019now}, comprising scans of real human subjects. We use these raw scans directly as partial inputs, a different protocol from the benchmark's official image-based challenge; NoW's reference scans, ground truth for that challenge, are not usable as shape-completion ground truth here, so we treat NoW as having none available and test generalization beyond the training distribution. \textbf{NPHM} is a second, independent real-scan dataset drawn from the NPHM head dataset~\cite{giebenhain2023nphm}: 271 identities, one scan each. NPHM's full-head scans do include hair, incompatible with our hairless UHM-based template, so we cannot evaluate on the full head or use it as shape-completion ground truth; instead we take a frontal crop matching NoW's framing, downsampled to 5k points, and evaluate the same way as NoW (\cref{sec:experiments:nphm}). Evaluation on both NoW and NPHM is limited to Chamfer distance (CD) as an accuracy proxy and qualitative inspection. CD is a surface-fit metric, not a shape-completion metric, a distinction \cref{sec:experiments:realscan} returns to. Every baseline deforms the same fixed-topology template, so a point's index is preserved through deformation, making both metrics directly comparable with no correspondence search.

\subsection{Baselines}
\label{sec:experiments:baselines}
We compare against six non-rigid registration methods spanning classical optimization, correspondence-free clustering, test-time optimization, and learned correspondence pruning: NRICP~\cite{amberg2007optimal}, CPD~\cite{myronenko2010cpd}, OAReg~\cite{zhao2025oareg}, ClusterReg~\cite{zhao2024clusterreg}, NSFP~\cite{li2021nsfp}, and GraphSCNet~\cite{qin2023graphscnet}. None predicts pose jointly with shape, so each requires a rigid pre-alignment stage before it can run at all. We give every one the same strong initialization, a 24-start octahedral multi-start point-to-point ICP, and report that cost as part of each baseline's time in \cref{tab:val,tab:now}. Ours predicts $(R,t)$ end to end in one forward pass and never invokes this or any external rigid-alignment stage, including during its own refinement (\cref{sec:method:refine}). We do not include iterative {3DMM}-fitting baselines: without a known or landmark-derived pose initialization, unavailable for an arbitrarily-posed raw scan, iterative fitting has no starting point short of the same external rigid-alignment stage our method avoids.

\subsection{Metrics}
\label{sec:experiments:metrics}
\textbf{Vertex error.} Direct per-vertex ground-truth error $\frac{1}{N_v}\sum_i \lVert \hat{\mathbf{M}}_i - \mathbf{M}^{\text{gt}}_i\rVert$, exploiting the shared template topology for exact correspondence (no nearest-neighbor search). We report the median over pairs within each crop category rather than the mean to mitigate any potential outlier biases across baselines.

\textbf{Chamfer distance (CD).} We report 1-way CD from the partial input scan to the deformed template, restricted to the visible region the scan actually covers (\cref{tab:val,tab:now}). CD re-searches nearest neighbors \emph{after} deformation, so a method that collapses or flattens the whole shape onto the visible region can still post a good CD by landing near some scan point post-hoc, independent of whether the rest of the shape is plausible. Rather than filtering methods on a separate structural-validity metric before reporting accuracy, we report GT error and CD side by side for every method in the same table (\cref{tab:val}): \cref{sec:experiments:results} shows several methods with competitive CD are simultaneously the worst by GT error, by an order of magnitude.

\subsection{Results: accuracy and Chamfer's blind spot}
\label{sec:experiments:results}

On the val set, ClusterReg and CPD, generic smoothness or correspondence-free regularizers with no face-specific prior, reach 38--47mm GT vertex error; ClusterReg in particular posts some of the lowest CD in the table despite this, while CPD's CD is merely competitive, not best-in-table. NRICP and OAReg show the same pattern less extremely. This is not four separate failures: inspecting the geometry directly (\cref{fig:teaser}) shows OAReg, ClusterReg, and CPD collapsing the head toward the visible region, which their regularizers have no mechanism to penalize, while CD only ever asks whether there is some nearby scan point. NSFP and GraphSCNet avoid outright collapse but still trail Ours by 3--4$\times$ on GT error, since their occluded-region behavior defaults to a generic guess rather than our learned, shape-specific completion (\cref{sec:method:regressor}). Ours leads every baseline on GT error, CD, and time simultaneously. \cref{sec:experiments:ablations} shows the GT-error and time advantages already hold for the forward pass alone, before refinement is applied.

\begin{table}[t]
  \centering
  \footnotesize
  \setlength{\tabcolsep}{3pt}
  \begin{tabular}{@{}lcc@{}}
    \toprule
    Method
      & CD (mm) $\downarrow$
      & Time (s) $\downarrow$ \\
    \midrule
    NRICP~\cite{amberg2007optimal}
      & $1.69 \,\pm\, 0.18$
      & 31.35 \\
    CPD~\cite{myronenko2010cpd}
      & $1.50 \,\pm\, 0.12$
      & 22.50 \\
    NSFP~\cite{li2021nsfp}
      & $0.61 \,\pm\, 0.05$
      & 11.80 \\
    GraphSCNet~\cite{qin2023graphscnet}
      & $1.61 \,\pm\, 0.09$
      & 37.45 \\
    ClusterReg~\cite{zhao2024clusterreg}
      & $1.05 \,\pm\, 0.06$
      & 23.05 \\
    OAReg~\cite{zhao2025oareg}
      & $1.61 \,\pm\, 0.14$
      & 10.90 \\
    Ours
      & $\mathbf{0.59 \,\pm\, 0.04}$
      & \textbf{2.26} \\
    \bottomrule
  \end{tabular}
  \caption{NoW benchmark test split, no ground truth: 1-way CD (partial scan$\to$deformed-template) and end-to-end wall-clock time, median $\pm$ std over the test scans, all methods timed on the same cluster hardware. CD uses each scan's own recovered scale factor rather than a fixed dataset-wide constant (\cref{sec:experiments:metrics}). Ours leads on both CD and time.}
  \label{tab:now}
  \vspace{6pt}
  \begin{tabular}{@{}lcc@{}}
    \toprule
    Method
      & CD (mm) $\downarrow$
      & Time (s) $\downarrow$ \\
    \midrule
    NRICP~\cite{amberg2007optimal}
      & $2.11 \,\pm\, 1.12$
      & 24.60 \\
    CPD~\cite{myronenko2010cpd}
      & $1.58 \,\pm\, 0.14$
      & 20.70 \\
    NSFP~\cite{li2021nsfp}
      & $0.80 \,\pm\, 0.11$
      & 13.60 \\
    GraphSCNet~\cite{qin2023graphscnet}
      & $1.68 \,\pm\, 0.35$
      & 20.10 \\
    ClusterReg~\cite{zhao2024clusterreg}
      & $1.14 \,\pm\, 0.09$
      & 25.40 \\
    OAReg~\cite{zhao2025oareg}
      & $1.56 \,\pm\, 0.34$
      & 12.60 \\
    Ours
      & $\mathbf{0.72 \,\pm\, 0.04}$
      & \textbf{2.08} \\
    \bottomrule
  \end{tabular}
  \caption{NPHM head dataset~\cite{giebenhain2023nphm}, face-only crops (full-head scans include hair, incompatible with our hairless template, so not usable as shape-completion ground truth): 1-way CD (partial scan$\to$deformed-template) and end-to-end wall-clock time, median $\pm$ std over 271 scans, all methods timed on the same cluster hardware (a different cluster from \cref{tab:val,tab:now}, so timings are comparable within this table but not across tables).}
  \label{tab:nphm}
\end{table}

\subsection{External dataset results}
\label{sec:experiments:realscan}
\label{sec:experiments:nphm}
At real-scan (NoW and NPHM) inference only, we scale $\sigma_d$ by a fixed $\times1.2$ correction factor (\cref{tab:config}): real scans have different point density than our synthetic training distribution even after unit-scale rescaling, and this factor was set by directly comparing real scans' and the template's point density during preprocessing.

On NoW, \cref{tab:now} shows Ours is simultaneously the most accurate and fastest method in the table: it leads NSFP, the strongest baseline, on CD (0.59mm vs.\ 0.61mm), in roughly a fifth of NSFP's time (2.26s vs.\ 11.80s), never paying NSFP's own rigid-ICP tax.

In Ours, the forward pass alone trades tight scan-fit for a regularized, morphable-model-consistent facial shape rather than directly optimizing surface fit, and posts a substantially worse raw CD without it (\cref{sec:experiments:ablations}). NSFP, by contrast, is a direct scan-fit optimizer initialized from an ICP pose, and wins CD for that reason regardless of whether its occluded-region completion is any good, which \cref{tab:val} shows it is not. Warm-starting the same objective from our own prediction closes this gap: the visible region snaps tightly to the scan surface, while the occluded region, which the refinement has no scan signal to correct, stays governed by our learned prior.

To check that our NoW results are not specific to that one benchmark, we repeat the real-scan evaluation on NPHM (\cref{sec:experiments:data}). \Cref{tab:nphm} shows the same pattern as \cref{tab:now}: Ours leads every baseline on CD while being the fastest method by a wide margin.

\begin{figure}[t]
  \centering
  \includegraphics[width=1.0\linewidth]{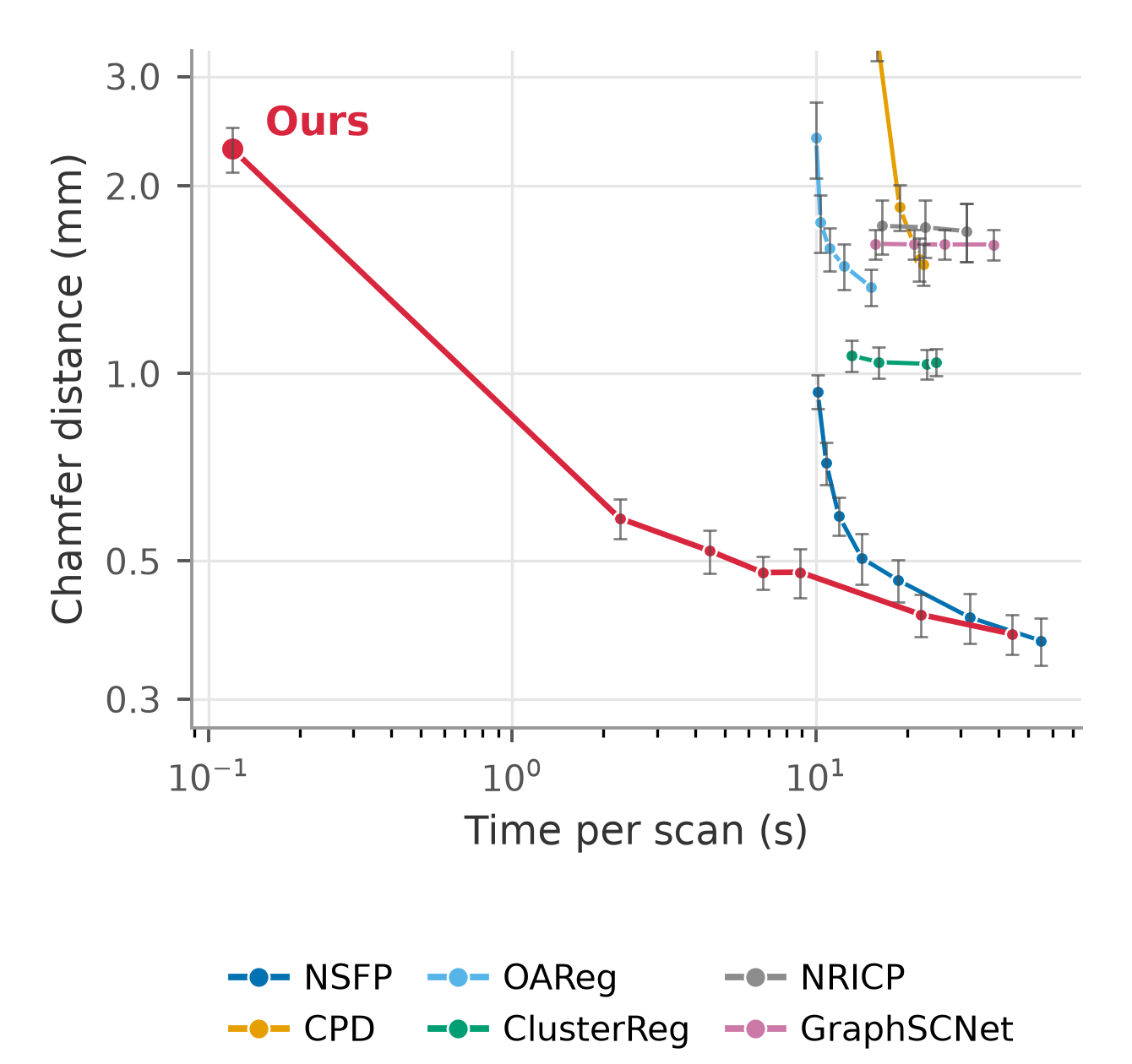}
  \caption{Accuracy (1-way CD, partial scan$\to$deformed-template, median) vs.\ wall-clock time (log scale, includes rigid-ICP pre-alignment for every baseline that requires it) across each method's own iteration/step budget, NoW benchmark test split. Baselines: NRICP~\cite{amberg2007optimal}, CPD~\cite{myronenko2010cpd}, NSFP~\cite{li2021nsfp}, GraphSCNet~\cite{qin2023graphscnet}, ClusterReg~\cite{zhao2024clusterreg}, OAReg~\cite{zhao2025oareg}.}
  \label{fig:pareto}
\end{figure}

\begin{figure*}[t]
  \centering
  \includegraphics[width=1.0\linewidth]{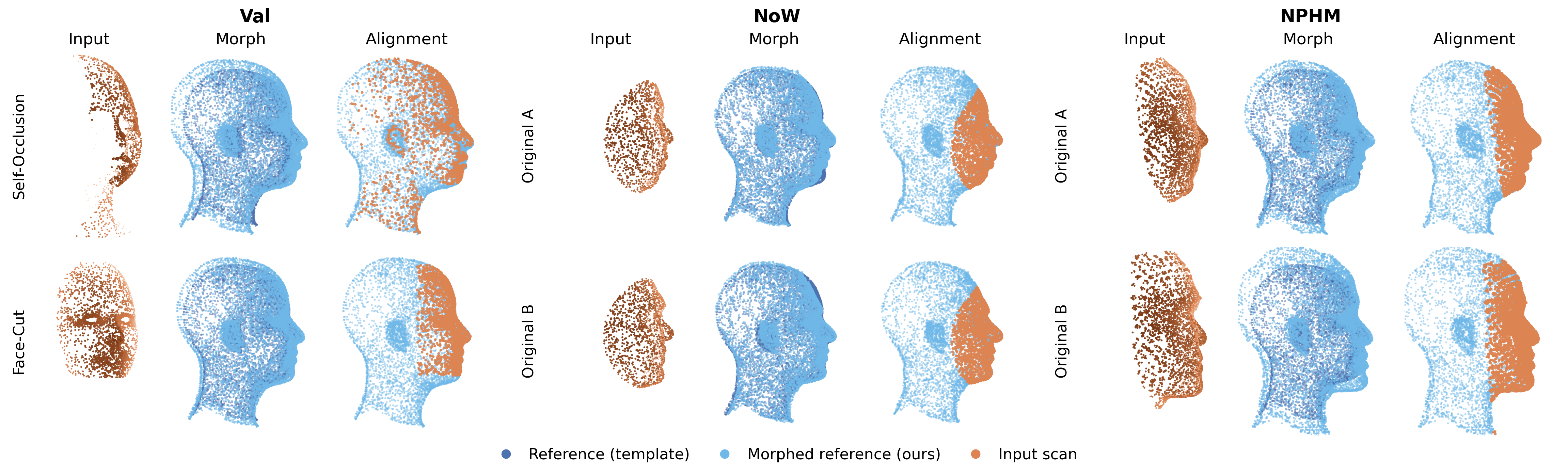}
  \caption{Qualitative results on Val (left), NoW (middle), and NPHM (right) examples. Rows: Self-Occlusion / Face-Cut crops (Val) and two representative scans each for NoW and NPHM. Columns: the partial input scan; our morphed reference alone, viewed from the side to show its departure from the fixed template (dark blue); and the same morphed reference (light blue) overlaid with the input scan (orange) to show surface-fit quality.}
  \label{fig:qualitative}
\end{figure*}

\subsection{Qualitative results}
\label{sec:experiments:qual}
\Cref{fig:qualitative} inspects individual Val, NoW, and NPHM examples directly. Each row is an independent example. The Morph column shows how far each row's morphed reference (light blue) departs from the fixed template (dark blue). The Alignment column overlays that same morphed reference with the row's input scan (orange) to show surface-fit quality.

\subsection{Ablation Studies}
\label{sec:experiments:ablations}

\textbf{Contribution of the refinement.} The forward pass alone, before refinement, already leads every baseline on GT error and by two to three orders of magnitude on time (2.71/2.13mm at 0.12s per scan on Val), confirming the accuracy gain in \cref{sec:experiments:results} comes from the learned completion itself, not the refinement. Its contribution shows up in CD instead. Without it, raw CD on NoW is 2.29mm, worse than every method in \cref{tab:now}, because the forward pass never fits the observed surface directly, by design (\cref{sec:method:refine}); adding it brings CD to 0.59mm (\cref{tab:now}) at a cost of roughly two additional seconds per scan.

\textbf{Accuracy vs.\ latency.} \Cref{fig:pareto} shows every method across its own step or iteration budget on NoW rather than a single operating point. Ours' single leftmost point is the forward pass alone, before any refinement steps, matching the 2.29mm raw CD reported above. From its cheapest refinement setting onward, Ours Pareto-dominates every baseline at every budget tested. Even 500 steps already beats every baseline's own highest-budget point except NSFP's, and pushing further, up to 10{,}000 steps, keeps improving CD with no plateau. This confirms \cref{tab:now}'s 500-step point is a genuine trade-off, not a ceiling, as more test-time compute keeps paying off but from a strictly better start.

\section{Discussion and limitations}
\label{sec:discussion}

\paragraph{Evaluating completion quality without ground truth.}
Ours closes the practical, scan-referenced accuracy gap on both NoW and NPHM (\cref{sec:experiments:realscan,sec:experiments:nphm,tab:now,tab:nphm}), but scan-referenced surface fit and occluded-region completion quality are different things, and we have no directly usable ground truth on either real-scan dataset to check the latter (NoW's own reference scans serve a different task; NPHM's full-head scans include hair, incompatible with our hairless template). Usable ground truth for NoW or NPHM would let those results be reported with the same confidence as our Val results.

\paragraph{Basis capacity and dataset scale.}
Our morphable prior is a linear PCA basis fit to 8{,}000 identities, patch-partitioned to keep local anatomical detail (\cref{sec:method:prelim}). This bias both enables the collapse-avoidance and accuracy in \cref{sec:experiments:results} and caps expressiveness at whatever the training population spans, so under-represented identities, expressions, or ages are unlikely to be well reconstructed. A learned, non-linear basis could relax this at the cost of the plausibility guarantee shown in \cref{tab:val}.

\paragraph{Broader impact.}
This work uses only point geometry, no color, texture, or identifying imagery, though improving recovery of complete facial geometry from partial scans is nonetheless dual-use, like biometric and face-reconstruction research generally. Training identities are synthetic, sampled from the published UHM model~\cite{ploumpis2020uhm} via the generation procedure of~\cite{foti2022swapdisentangled}, not collected from real subjects; both are used under their published terms, as are NoW~\cite{sanyal2019now} and NPHM~\cite{giebenhain2023nphm}. We see no immediate deployment risk here.

\section{Conclusion}
\label{sec:conclusion}

We presented a single network that jointly predicts a morphable-model-based facial-shape completion and rigid pose for partial {3D} face scans, closing with a short scan-referenced refinement. Our evaluation protocol reports per-vertex ground-truth error alongside Chamfer distance, exposing a collapse failure mode Chamfer distance alone conceals. Our method avoids this failure, leading held-out Val identities by 3--4$\times$ over the strongest baselines (\cref{tab:val}) and both NoW's and NPHM's ground-truth-free scans in accuracy and speed (\cref{tab:now,tab:nphm}).

\newpage
{
    \small
    \bibliographystyle{ieeenat_fullname}
    \bibliography{main}
}

\end{document}